\documentclass[letterpaper, 10 pt, conference]{ieeeconf}  
\IEEEoverridecommandlockouts                              

\usepackage{graphics} 
\usepackage{epsfig} 
\usepackage{times} 
\usepackage{amsmath, amsfonts} 
\usepackage{amssymb}  
\usepackage{bm}

\usepackage{subcaption}
\usepackage{algorithmicx}
\usepackage{algorithm}%
\usepackage{algorithmicx}%
\usepackage{algpseudocode}%
\usepackage{stfloats}
\usepackage{hyperref}
\usepackage[font={small}]{caption}

\usepackage[nolist,nohyperlinks]{acronym}
\acrodef{GPS}[GPS]{Global Positioning System}
\acrodef{SLAM}[SLAM]{Simultaneous Localization And Mapping}
\acrodef{GPS}[GPS]{Global Positioning System}
\acrodef{RTK}[RTK]{Real-time Kinematics}
\acrodef{GNSS}[GNSS]{Global Navigation Satellite System}
\acrodef{ROS}[ROS]{Robot Operating System}
\acrodef{API}[API]{Application Programming Interface}
\acrodef{UAV}[UAV]{Unmanned Aerial Vehicle}
\acrodef{MAV}[MAV]{Micro Aerial Vehicle}
\acrodef{UGV}[UGV]{Unmanned Ground Vehicle}
\acrodef{IMU}[IMU]{Inertial Measurement Unit}
\acrodef{MPC}[MPC]{Model Predictive Control}
\acrodef{LiDAR}[LiDAR]{Light Detection and Ranging}
\acrodef{ESC}[ESC]{Electronic Speed Controller}
\acrodef{LKF}[LKF]{Linear Kalman Filter}
\acrodef{UKF}[UKF]{Unscented Kalman Filter}
\acrodef{EKF}[EKF]{Extended Kalman Filter}
\acrodef{RAS}[RAS]{Robotics and Automation Society}
\acrodef{IEEE}[IEEE]{Institute of Electrical and Electronics Engineers}
\acrodef{MRS}[MRS]{Multi-robot Systems Group}
\acrodef{CdTe}[CdTe]{Cadmium Telluride}
\acrodef{RMSE}[RMSE]{Root Mean Square Error}
\acrodef{FDNPP}[FDNPP]{Fukushima Daiichi Nuclear Power Plant}
\acrodef{MLEM}[MLEM]{Maximum Likelihood Expectation Maximization}
\acrodef{EM}[EM]{Expectation Maximization}
\acrodef{TSP}[TSP]{Travelling salesman problem}
\acrodef{pix}[Minipix]{MiniPIX TPX3}
\acrodef{CC}[CC]{Compton camera}

\newcommand{\red}[1]{\textcolor{red}{#1}}

\newcommand{\blue}[1]{\textcolor{blue}{#1}}
\newcommand{\green}[1]{\textcolor{green}{#1}}

\newcommand{\appropto}{\mathrel{\vcenter{
  \offinterlineskip\halign{\hfil$##$\cr
    \propto\cr\noalign{\kern2pt}\sim\cr\noalign{\kern-2pt}}}}}
\newcommand{\reffig}[1]{Figure~\ref{#1}}

\newcommand{\refalg}[1]{Alg.~\ref{#1}}
\newcommand{\refsec}[1]{Section~\ref{#1}}
\newcommand{\reftab}[1]{Table~\ref{#1}}
\newcommand{\refeq}[1]{\eqref{#1}}
\usepackage{siunitx}
\DeclareSIUnit \parsec {pc}
\DeclareSIUnit \electronvolt {eV}
\DeclareSIUnit \pixel {px}
\DeclareSIUnit \arcmin {arcmin}
\DeclareSIUnit \erg {erg}
\DeclareSIUnit \joul {J}

\title{\LARGE \bf
Autonomous localization of multiple ionizing radiation sources using miniature single-layer Compton cameras onboard a group of micro aerial vehicles
}
\author{Michal Werner$^1$, Tom\'{a}\v{s} B\'{a}\v{c}a$^{1}$, Petr \v{S}tibinger$^{1}$,\\ Daniela Doubravov\'{a}$^{2}$, Jaroslav \v{S}olc$^{3}$, Jan Rus\v{n}\'{a}k$^{3}$, and Martin Saska$^{1}$
\thanks{$^{1}$Authors are with the Faculty of Electrical Engineering, Czech Technical University in Prague, Technick\'{a} 2, Prague 6, {\tt\footnotesize wernemic@fel.cvut.cz}.}%
\thanks{$^{2}$ Authors are with Advacam s.r.o., U Pergamenky 12, 170 00, Prague.}
\thanks{$^{3}$ Authors are with Czech Metrology Institute, Okruzni 31, 638 00, Brno.}
}

\begin{document}

\maketitle
\thispagestyle{empty}
\pagestyle{empty}

\begin{abstract}
A novel method for autonomous localization of multiple sources of gamma radiation using a group of \acp{MAV} is presented in this paper.
The method utilizes an extremely lightweight (44\,g) Compton camera \acl{pix}.
The compact size of the detector allows for deployment onboard safe and agile small-scale \acp{UAV}.
The proposed radiation mapping approach fuses measurements from multiple distributed Compton camera sensors to accurately estimate the positions of multiple radioactive sources in real time.
Unlike commonly used intensity-based detectors, the Compton camera reconstructs the set of possible directions towards a radiation source from just a single ionizing particle.
Therefore, the proposed approach can localize radiation sources without having to estimate the gradient of a radiation field or contour lines, which require longer measurements.
The instant estimation is able to fully exploit the potential of highly mobile \acp{MAV}.
The radiation mapping method is combined with an active search strategy, which coordinates the future actions of the \acp{MAV} in order to improve the quality of the estimate of the sources' positions, as well as to explore the area of interest faster. 
The proposed solution is evaluated in simulation and real-world experiments with multiple Cesium-137 radiation sources. 
\end{abstract}


\section{INTRODUCTION}
The fast localization of sources of ionizing radiation is a critical task for public safety.
\acp{UAV} equipped with the appropriate sensors present a viable solution for such a task.
The \acp{UAV} are able to not only speed up the detection process, but to replace human workers in hazardous environments, thereby limiting the risk of exposure to harmful radiation imperceptible by human senses.
Recent advancements in the field of mobile robotics and sensory equipment have opened new possibilities in the remote sensing of ionizing radiation.
We present a radiation mapping method that exploits the benefits and operability of small-scale \acp{UAV} equipped with miniature direction-based radiation detectors.

Remote sensing of ionizing radiation is typically done using intensity-based detectors.
The use of sensors measuring particle flux at a given position requires scanning the entire area \cite{sanada_2015_big_helicopter_fukushima_intensity}, estimating the contour lines \cite{redwan_newaz_2016_contour} or the gradient \cite{mascarich_autonomous_2021_ICRA} of the radiation field.
The stochastic nature of nuclear fission makes the dosimetric measurements noisy, which needs to be compensated for either by the size of the detector (increased sensitivity) or by acquiring the measurements for a sufficiently long time.
Highly sensitive detectors weighing several kilograms must be carried by platforms with a sufficient payload capacity \cite{sanada_2015_big_helicopter_fukushima_intensity,towler_2012_helicopter_contour}, which limits their operability as they are not able to fly close to obstacles, humans, or in cluttered environments.
Using miniaturized dosimeters requires longer acquisition times (to record a sufficient number of interactions), which slows down the movement of the \ac{UAV} and does not fully exploit its potential.

\begin{figure}[!htb]
  \centering
  \includegraphics[width=0.48\textwidth,trim={0cm 2.9cm 0cm 4cm},clip]{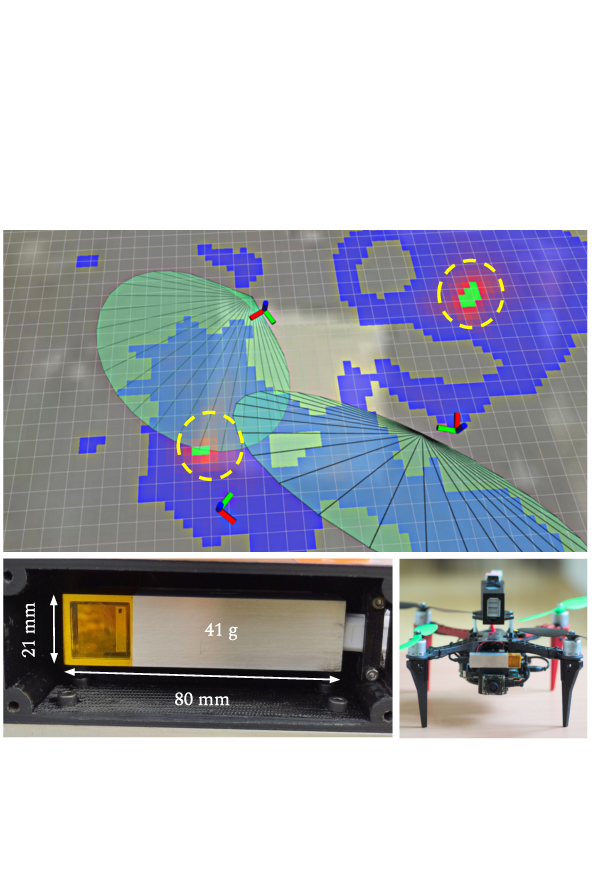}
  \label{e1:lam}
  \caption{Estimate of the radioactive hotspots (yellow circles) generated from the Compton measurements (green cones) by the proposed method in a real-world experiment (top). The estimation method builds on a small and lightweight Compton camera (bottom left) that can be deployed onboard sub-1 kg \acp{MAV} with onboard processing (bottom right).}
  \label{fig:}
\end{figure}

Our approach builds on the miniature Compton camera, which belongs to the direction-based class of detectors.
A single gamma particle interacting with the matter of the detector in the form of the Compton scattering effect \cite{compton} allows us to reconstruct a set of possible directions towards the radioactive source.
The event-based nature of the measurements is beneficial, as the \ac{MAV} is able to keep flying at an arbitrary speed without needing to stop or otherwise constrain its motion.
We present a method that fuses the Compton measurements in order to identify multiple sources of ionizing radiation located within the area.
We utilize the lightweight and compact \ac{pix} Compton camera (weighing only $\SI{44}{\gram}$ with dimensions $80 \times 21 \times 14 \ \si{\milli \meter}$). 
Thanks to its compactness, the \ac{pix} Compton camera can be mounted onboard agile and lightweight \acp{MAV}.
The lower effective volume of gamma-absorbent material in the miniature detector
is compensated for, as the small and lightweight \acp{MAV} are able to fly closer to possible radiation sources when compared to bulky \acp{UAV} or unmanned helicopters.
In contrast to intensity-based detectors, the speed of the \ac{MAV} does not influence the quality of measurements obtained by the onboard Compton event camera.
We can, therefore, fully exploit the potential of small and agile \acp{MAV} capable of reacting to the newly obtained measurements.
Furthermore, the localization process is sped up by the deployment of a team of \ac{MAV} simultaneously measuring in different locations and the coordination of their actions.
We present a novel solution for onboard real-time Compton-camera-based radiation mapping, combined with an active search strategy that guides the \acp{MAV}' future actions based on previous measurements.
The method was evaluated in simulation, as well as in real-world experiments.

\subsection{Related work}
The need for efficient methods for sensing ionizing radiation emerged following the earthquake and subsequent disaster at the Fukushima Daiichi Nuclear Power Plant in 2011.
The contaminated area has since been explored using both ground \cite{sato_radiation_2019_robot_compton, towler_2012_ground_robots_fukushima} and aerial robots \cite{sanada_2015_big_helicopter_fukushima_intensity, towler_2012_helicopter_contour, jiang_2016_helicopter_compton}.
In consideration to the related literature, it is evident that ground robots offer a larger payload, although aerial robots offer enhanced mobility. Additionally, most existing state-of-the-art approaches rely on intensity-based detectors.
An unmanned helicopter equipped with a large scintillator measuring radiation intensity at every position of the predefined path was presented in \cite{sanada_2015_big_helicopter_fukushima_intensity}.
Martin et al. \cite{martin_2016_one_drone_intensity} showed intensity measurements along a predefined path combined with a 3D topological map.
Both \cite{sanada_2015_big_helicopter_fukushima_intensity} and \cite{martin_2016_one_drone_intensity} required scanning the whole area, which is a time-consuming and less efficient procedure, as the predefined path does not take into account past measurements and the drone spends less time in the vicinity of the radiation sources.
The estimation of a radiation field's contour lines was presented in  \cite{towler_2012_helicopter_contour, han_2013_contour_mapping}, which required measurements at many different positions.
A solution for indoor radiation mapping in a GPS-denied environment was presented in \cite{mascarich_autonomous_2021_ICRA}. Therein, the intensity-based approach relied on estimating the gradient of the radiation field based on a sufficient number of measurements acquired by miniature scintillators at multiple positions.
Additionally, the drone was able to plan its future movements to improve the estimation quality.
The gradient estimation from neighbouring cells assumes a sufficient number of intensity measurements, which constrains the movement of the \ac{UAV}.

A Compton camera as a direction-based radiation sensor was utilized in the following contributions.
Sato et al. \cite{sato_2019_robot_fukushima_compton_single_axis} presented a large multi-layer Compton camera mounted on a crawler robot moving in one direction.
Outdoor radiation imaging using a two-layer Compton camera weighing $\SI{1.5}{\kilogram}$ mounted on a drone following a predefined path was presented in \cite{sato_2020_compton_drone}. 
Therein, the Compton measurements were projected onto a 3D map, but all measurements were processed offline.
Cong et al. \cite{cong_2020_compton_robot} showed a single source localization method in 3D for Compton measurements based on the back projection of cones in the image space.
The \acf{MLEM} method for Compton measurements reconstruction was originally developed for nuclear imaging in medicine \cite{ wilderman_1998_list_mode_mlem}. 
Authors of \cite{kim_2017_robot_compton} presented 
an offline localization of a single simulated source using the \ac{MLEM} method for a ground robot equipped with a two-layer Compton camera.
However, the authors do not model the sensitivity of detection, which is crucial for the accurate autonomous localization of multiple sources distributed in the area.
Vettel et al. \cite{vetter_2018_compton_mlem} presented 3D source localization using the \ac{MLEM} method and a multilayer Compton camera weighing several kilograms.
The search space was restricted to the surface of obstacles obtained by the 3D mapping method. 
Unlike our solution, all the data were processed offline.

Methods for single radiation source localization utilizing the same \ac{pix} radiation detector were presented in \cite{stibinger_2020_ral} and \cite{baca_gamma_2021_icuas}.
In \cite{stibinger_2020_ral}, the drone estimates the vector towards the radiation source by measuring intensity (number of recorded events) in all directions while rotating around the vertical axis.
The measurements are then fused by a \ac{LKF}.
The approach in \cite{stibinger_2020_ral} is time-consuming, since the drone must hover at a fixed position while obtaining the measurements.
In \cite{baca_gamma_2021_icuas}, the authors fuse the \ac{pix} Compton camera measurements with a \ac{LKF} to localize a single (static or moving) source.
Neither \cite{stibinger_2020_ral} nor \cite{baca_gamma_2021_icuas} can be used for the localization of multiple radiation sources.
Compared to all of the works mentioned above, we present a solution that can localize any number of compact or distributed radiation sources by deploying multiple agile and small \acp{MAV} equipped with lightweight and compact Compton cameras.

\subsection{Contribution}
This work presents a novel method for the autonomous localization of multiple ionizing radiation sources by a group of \acp{MAV} carrying miniature \ac{pix} Compton cameras onboard.
In particular, we solve the task of online radiation mapping from Compton measurements by utilizing the maximum-likelihood method and modeling the statistical properties of the single-layer Compton camera (\refsec{sec:modeling_the_sensor_properties}).
We introduce a simplified model for cone projection (\refsec{sec:system_matrix}) and a memory-efficient representation of detection sensitivity (\refsec{sec:sensitivity}).
Therefore, we make the reconstruction method tractable for onboard online estimation in real-world scenarios.  
As a minor contribution, we combine the radiation mapping method with an active search strategy (\refsec{sec:search_strategy}) and demonstrate in both simulation and real-world experiments. 

\section{PROBLEM DESCRIPTION}
\subsection{Challenges}
\label{sec:challenges}
The localization of multiple radioactive sources using Compton measurements is a challenging task due to 
several phenomena stemming from nuclear and particle physics.
The key factors are:
\textbf{A) The inverse square law}: the intensity of radiation decreases with $\approx \frac{1}{d^2}$ as the distance from the source $d$ grows, which limits the sensing range.
Therefore, obtaining accurate measurements with a miniature radiation detector requires flying as close as possible to the radiation source.
\textbf{B) An unknown number of radiation sources}: it is impossible to decide whether a particular Compton measurement originated from an already localized source of ionizing radiation. 
\textbf{C) Noise in measurements, background radiation}: the measurements may contain outliers, and the single-layer Compton camera may produce multiple hypotheses from a single recorded particle.
\textbf{D) Measurement-trajectory dependency}: assuming 1) an arbitrary trajectory during the search, 2) the inverse square law, and 3) the presence of an unknown number of radiation sources with unknown (non-uniform) emission activity, we must consider the past movement of the drone in the estimation method.
Additionally, the fact that the \ac{MAV} (equipped with a radiation detector) did not measure any ionizing particles in some areas should also be incorporated. Thus, modeling the coverage of the area is beneficial to the active search strategy.
The proposed method addresses these factors for distributed Compton imaging using mobile \acp{MAV}.

\subsection{Compton camera}
\begin{figure}[ht!]
    \centering
    \begin{subfigure}{0.22\textwidth}
        \includegraphics[width=\textwidth]{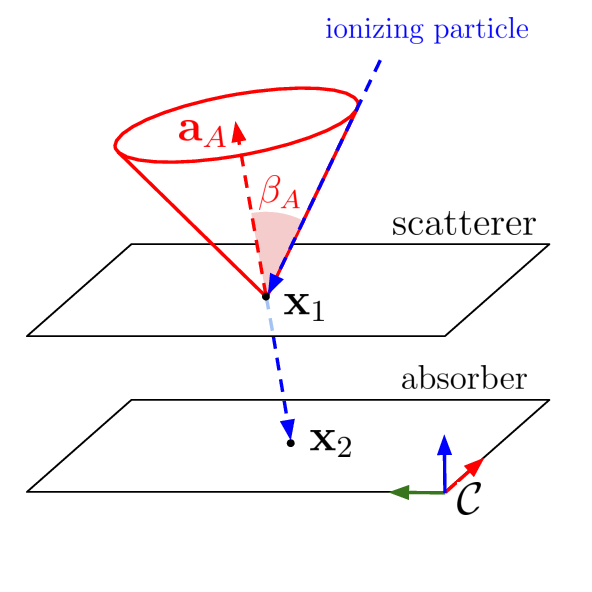}
        \caption{Classical Compton camera}
        \label{fig:classic_cc}
    \end{subfigure}
    \begin{subfigure}{0.22\textwidth}
        \includegraphics[width=\textwidth]{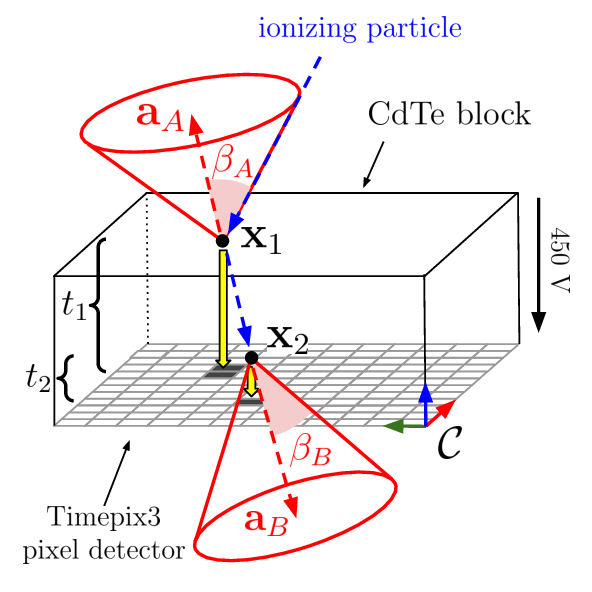}
        \caption{Minipix Timepix3}
        \label{fig:timepix_cc}
    \end{subfigure}
    \caption{ Differences between the classical two-layer Compton camera architecture and the single-layer Minipix Timepix3.}
    \vspace{-0.7cm}
\end{figure}

The Compton camera uses the Compton scattering effect \cite{compton} to reconstruct the set of possible directions from which the ionizing particle 
could have reached the sensor.
The Compton camera is typically composed of two detectors: a \textit{scatterer} and an \textit{absorber}.
The incident photon with energy $E_{0}$ first interacts with the \textit{scatterer} 
where the Compton scattering occurs.
An electron with energy $E_{1}$ is a bi-product of the interaction, which is immediately captured by the \textit{scatterer}, and its position $\mathbf{x}_{1}$ and measured energy are recorded.
A second product of the interaction, a lower energetic photon with energy $E_{2}$, is scattered at a (Compton) angle $\beta$.
The scattered photon is then absorbed by the \textit{absorber}.
The absorbed energy $E_{2}$ and the position of the interaction $\mathbf{x}_{2}$ are measured and recorded.
According to \cite{compton}, the scattering angle $\beta$ can be reconstructed (following \cite{baca_2019_timepix_iros}) as:
\begin{equation}
  \beta = \mathrm{cos}^{-1} 
  \underset{B}{\underbrace{\left (
   1+m_{e}c^{2} \left( \frac{1}{E_{1}+E_{0}} - \frac{1}{E_{0}}\right )  \right )
  }},
  \label{eq:compton_beta_formula}
\end{equation}
where $m_{e}$ is is the electron rest mass, $c$ is the speed of light in a vacuum, and $0<B<1$.
Since the Compton scattering is a symmetrical phenomenon, the set of possible directions of incoming particles forms the surface of a cone.
The conical surface (denoted as a  Compton cone) is parametrized by the cone axis $\mathbf{a} = \mathbf{x}_{2}-\mathbf{x}_{1}$, the Compton scattering angle $\beta$, and the origin of the cone.
The geometry is illustrated in \reffig{fig:classic_cc}.

\subsection{Minipix Timepix3}
The \ac{pix} detector is composed of a Timepix3 pixel detector \cite{poikela_2014_timepix},  a sensor body made of a compact block of \ac{CdTe} semiconductor material (with dimensions $14 \times 14 \times 2 \ \si{\milli\meter}$), and the Minipix readout electronics.
Unlike other types of detectors, the event-based \ac{pix} sensor can report the 
detected $\gamma$ particles in almost real time. This allows us to use it for an active strategy where autonomous \ac{UAV}s react 
to the measurements acquired during the flight.
The incoming ionizing radiation interacts with the sensor's matter and separates the electric charge from the \ac{CdTe} material.
The separated electrons are accelerated by a $\SI{450}{\volt}$ electric potential towards one facet of the Timepix3 pixel detector.
Given the measured arrival times and energy of interactions, the coinciding products of Compton scattering might be paired together, assuming that both the Compton scattering and the follow-up photon absorption happened simultaneously.
More technical details related to the sensor operation are provided in \cite{baca_2019_timepix_iros}.
However, the matched pair of events recorded by the pixel detector results in two possible hypotheses.
As shown in \reffig{fig:timepix_cc}, a single ionizing particle produces two Compton cones $A$ and $B$.
Unlike the two-layer Compton cameras, the unshielded \ac{pix} records incoming particles from all directions and does not have a limited field of view.
The maximum likelihood approach can handle the ambiguity of measurements, as up to two Compton cones are produced by a single ionizing particle.

\section{RADIATION MAPPING METHOD}

\subsection{Preliminaries}
\begin{figure}[!htb]
  \centering
  \includegraphics[width=0.45\textwidth,trim={0cm 0cm 0cm 0.6cm},clip]{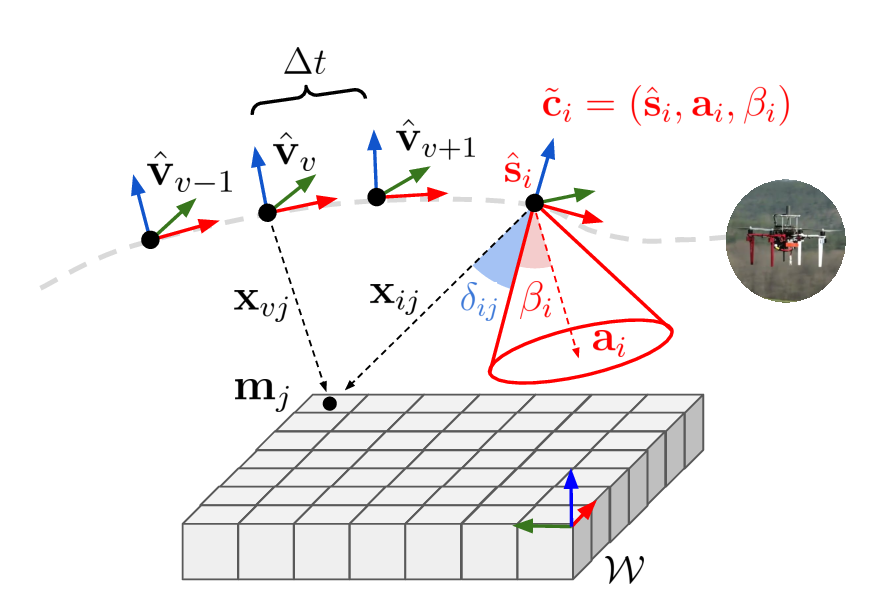}
  \caption{Compton camera produces Compton cones $\tilde{\mathbf{c}}_{i} = (\mathbf{\hat{s}}_{i}, \mathbf{a}_{i}, \beta_{i})$ parameterized by the cones' origin, axis, and the Compton angle, respectively. As the \ac{MAV} flies through the environment, the viewpoints $\mathbf{\hat{v}}$ are sampled. The area of possible source positions $\mathcal{M}$ is discretized into $J$ cells, where each cell is represented by its center position $\mathbf{m}_{j}$.}
  \label{fig:setup}
  \vspace{-0.7cm}
\end{figure}

\begin{table}[!htb]
  \centering
  \begin{tabular}{lll}
    $\mathbf{x}$, $\lVert \mathbf{x}\rVert$ & vector in 3D, Euclidean norm of $\mathbf{x}$ \\
    $\mathbf{\hat{s}} = (\mathbf{s},\alpha_{0}, \alpha_{1}, \alpha_{2})$ & oriented sensor pose in 3D \\
    $\tilde{\mathbf{c}}_{i} = (\mathbf{\hat{s}}_{i}, \mathbf{a}_{i}, \beta_{i})$ & Compton cone with sensor pose $\mathbf{\hat{s}}$,\\ & cone-axis vector $\mathbf{a}$, and scattering angle $\beta$ \\
    $\mathbf{m}_{j} \in \mathcal{M}$ & map position with index $j \in \{1, \dots, J \} $ \\
    $\mathbf{T} \in \mathbb{R}^{I \times J}$, $\mathbf{s} \in \mathbb{R}^{J}$  & system matrix, sensitivity vector \\
    $\bm{\lambda} \in \mathbb{R}^{J}$ & vector of hidden parameters \\
  \end{tabular}
  \caption{Mathematical notation and notable symbols.}
  \label{tab:mathematical_notation}
\end{table}
As the \acp{UAV} fly through the environment, each recorded Compton camera measurement $\tilde{\mathbf{c}}_{i} = (\mathbf{\hat{s}}_{i}, \mathbf{a}_{i}, \beta_{i})$ is parameterized by the sensor pose $\mathbf{\hat{s}}$ at the time the measurement was taken, the cone-axis vector $\mathbf{a}$ and the scattering angle $\beta$.
The Compton cones are indexed with $i\in \{ 1, \dots, I\}$, where $I$ is the number of Compton cones recorded by all the \acp{UAV} until the current time.
The trajectories of the \acp{UAV} are sampled with a time period $\Delta t$ producing viewpoints $\mathbf{\hat{v}}_{v}$, where $v \in \{1, \dots, V\}$ and $V$ denotes the total number of viewpoints collected until the current time. 
The area of interest, where the sources of ionizing radiation might be located, is discretized into $J$ discrete bins with resolution $r$.
Each of the bins is represented by its center position $\mathbf{m}_{j}$ and indexed with $j \in \{1, \dots, J\}$, as illustrated in \reffig{fig:setup}.
In this paper, we assume that the sources of ionizing radiation are located somewhere on the 
terrain's surface.
Any voxelized 3D map representation (such as an Octomap) can be used as the set $\mathcal{M}$.

\subsection{Maximum likelihood estimation}
For designing the radiation mapping approach, we take inspiration from the \acf{MLEM} iterative estimation method, which was originally proposed for nuclear medicine applications \cite{wilderman_1998_list_mode_mlem}.
We assume that the number of ionizing particles emitted from the position $\mathbf{m}_{j}$ is a discrete random variable following a Poisson distribution with the expected value $\lambda_{j}$.
Every map cell $\mathbf{m}_{j}$ is a possible radiation source and, therefore, the vector of the unknown parameters $\bm{\lambda}\in \mathbb{R}^{J}$ to be estimated is defined as $\bm{\lambda} = \left [ \lambda_{1}, \dots, \lambda_{J} \right ]$.
We further define the matrix $\mathbf{T} \in \mathbb{R}^{I \times J}$ (referred to in the literature as the system matrix) with elements
\begin{equation}
  t_{ij} =  P(\textrm{detected in } i | \textrm{emitted from } j).
  \label{eq:tij}
\end{equation}
In other words, $t_{ij}$ represents the probability that we observe the measurement $i$ given that a radioactive particle that caused the measurement $i$ has been emitted from position $\mathbf{m}_{j}$.
Since the measured data are not complete (not every emitted particle reaches the detector and is detected), we define the sensitivity of detection $\mathbf{s}\in \mathbb{R}^{J}$ with the elements:
\begin{equation}
  s_{j} =  P(\textrm{detected} | \textrm{emitted from } j),
  \label{eq:sj}
\end{equation} 
which models the chance that a particle emitted from $j$ is detected by any of the detectors onboard the \acp{MAV}.
In other words, the sensitivity of detection can be interpreted as the coverage of the area (which is useful for the active search strategy) and enables the accurate localization of multiple radiation sources with different emission activities.   

The \ac{MLEM} algorithm estimates the $\bm{\lambda}$ at every map position $j$ by maximizing the likelihood of measured data.
The likelihood cannot be maximized directly. Therefore, it utilizes the \ac{EM} algorithm \cite{EM}.
One iteration of the \ac{MLEM} algorithm is defined as 
\begin{equation}
  \lambda_{j}^{[l]} = \frac{\lambda_{j}^{[l-1]}}{s_{j}} \sum_{i \in \{1,\dots,I\}} \frac{t_{ij}}{\sum_{k \in \{1,\dots,I\}} t_{ik} \lambda_{k}^{[l-1]}},
  \label{eq:MLEM}
\end{equation}
where $l$ is the current iteration, $\lambda_{j}$ is the unknown hidden parameter to be estimated, $t_{ij}$ is an element of the system matrix, and $s_{j}$ is the sensitivity of detection.
To use the \ac{MLEM} method, we need to model the sensitivity vector $\mathbf{s}$ and system matrix $\mathbf{T}$, which is dependent on the particular sensor device and application.
The computation speed is the key factor for running the \ac{MLEM} algorithm online during the search.
Unlike in medical applications where the source-detector distance is low ($\leq 1\,m$), data is processed offline, and the number of measurements is high, we need to make reasonable simplifications to speed up the reconstruction process.

\subsection{Modeling the sensor properties}
\label{sec:modeling_the_sensor_properties}
\begin{figure}[ht!]
    \centering
    \begin{subfigure}{0.18\textwidth}
        \includegraphics[width=\textwidth]{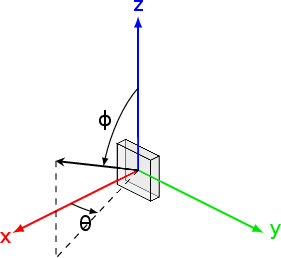}
        \caption{The geometry of the detector and polar coordinates}
        \label{fig:detector_axis}
    \end{subfigure}
    \begin{subfigure}{0.26\textwidth}
        \includegraphics[width=\textwidth,trim={0cm 2cm 0cm 1cm}]{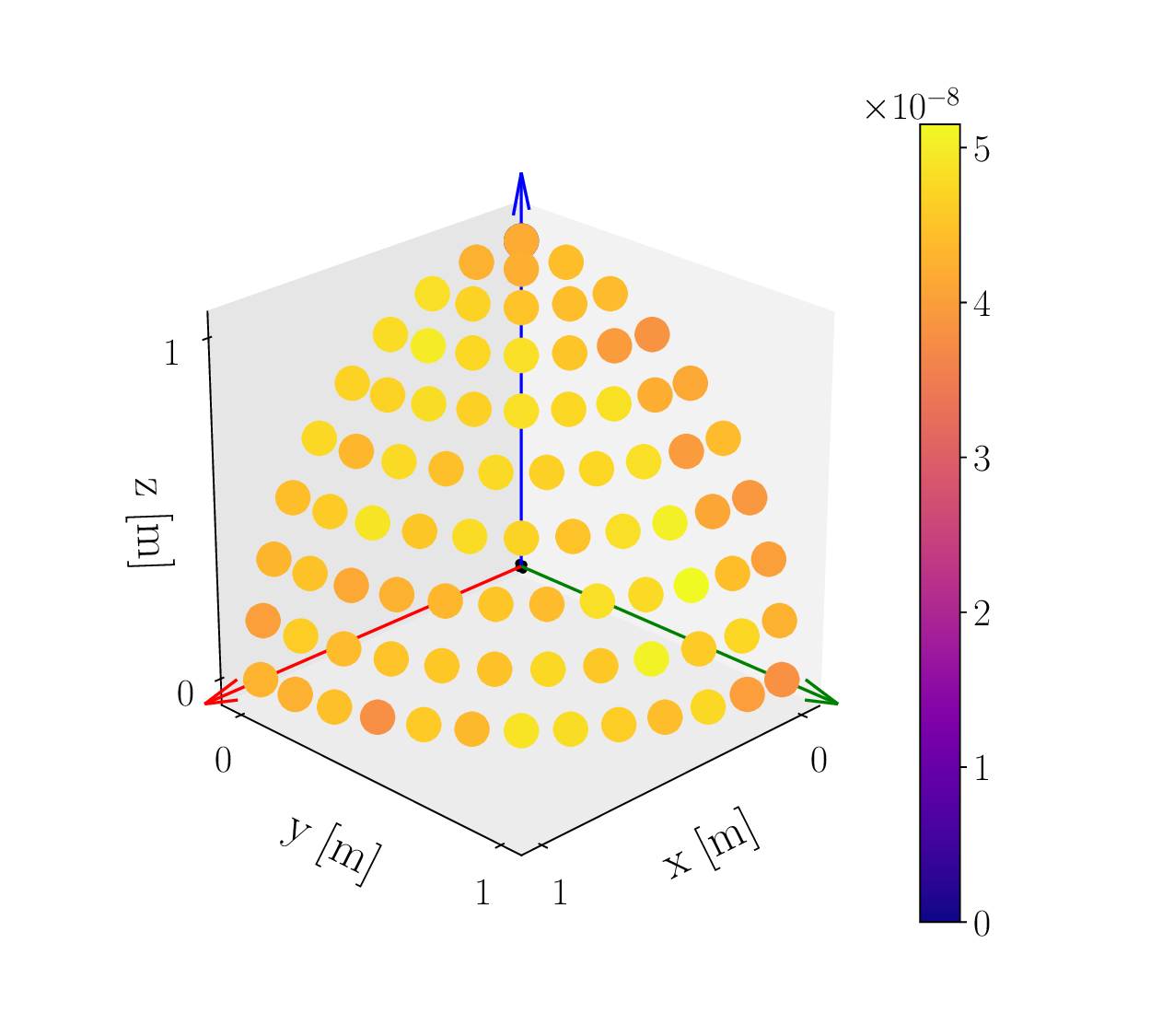}
        \caption{p(cone) for Cs-137}
        \label{fig:mc}
    \end{subfigure}
    \caption{Monte Carlo simulation showing the probability of Compton detections at various sensor orientations (i.e., relative positions of the source) with respect to the detector's geometry.}
    \vspace{-0.7cm}
\end{figure}

The probabilistic approach requires modeling equations \refeq{eq:tij} and \refeq{eq:sj}.
The probability of Compton scattering and other particle interactions largely depends on the intersection length between the particle trajectory and the \ac{CdTe} block.
Using a Monte Carlo simulation, we model the probability of 
a Compton measurement caused by a particle approaching the \ac{pix} detector from a certain direction. 
Although the analytical methods for multi-layer Compton cameras exist, it is difficult to model all the underlying particle interactions inside the Compton camera accurately, as well as unnecessary for the given application.
The simulated sources are placed on a sphere with a radius of $\SI{1}{\meter}$ around the detector (the air attenuation is not taken into consideration).
Each source position is parameterized with polar angles $\phi, \theta$ as shown in \reffig{fig:detector_axis}.
Every simulated source emits $10^7$ particles in all directions.
We count the particles emitted from the simulated source towards the visible detector surface that causes interactions resulting in successful Compton measurements.
The details of the simulator can be found in \cite{baca_2019_timepix_iros}.
The estimated probability is stored in a lookup table $\mathrm{L}(\phi, \theta)$.
For any pair of polar angles $\phi_{querry}, \theta_{qeurry}$, the lookup table returns the stored probability of Compton scattering associated with the nearest $\phi, \theta$ precomputed pair. 
The simulation results for Cesium-137 are illustrated in \reffig{fig:mc}.
We notice that the direction sensitivity is nearly uniform w.r.t. different angles for Cesium-137, despite the uneven geometry of the detector.
Although the particle emitted towards the side of the detector ($\theta \approx \frac{\pi}{2}$) has a lower chance of hitting the detector surface when compared to the particle emitted towards the front of the detector $\phi \approx \frac{\pi}{2}, \theta \approx 0$, the longer intersection with the \ac{CdTe} block increases the chance of a Compton measurement.

\subsection{Sensitivity estimation}
\label{sec:sensitivity}
The sensitivity of the detection $s_{j}$ estimates the chance that a particle emitted at position $\mathbf{m}_{j}$ has been recorded by the Compton camera onboard the \acp{MAV}.
First, we model the sensitivity of detection for the viewpoint $\mathbf{\hat{v}}_{v}$ and map position $\mathbf{m}_{j}$ pair as:
\begin{equation}
      s_{jv} = e^{-(\mu \lVert \mathbf{m}_{j} - \mathbf{v}_{v} \rVert)}\ 
  \frac{\mathrm{L}(\phi_{jv}, \theta_{jv})}{\lVert \mathbf{m}_{j} - \mathbf{v}_{v} \rVert ^ {2} },
\end{equation}
where $\mu$ is the linear attenuation coefficient ($\approx \SI{0.01}{\meter^{-1}} $ for $\SI{622}{\kilo\electronvolt}$ photons in the air), 
$\phi_{jv}, \theta_{jv}$ are the polar coordinates determining the relative position of $\mathbf{m}_{j}$, and the sensor position $\mathbf{\hat{v}}_{v}$. 
The sensitivity of detection $s^{[t+1]}$ at time $[t+1]$ is computed as follows:
\begin{equation}
  s_{j}^{[t+1]} = s_{j}^{[t]} + \sum_{v \in V^{[t:t+1]}} s_{jv} \Delta_{v}, 
  \label{eq:sen_iter}
\end{equation}
where the sum $\sum_{v \in V^{[t:t+1]}}$ iterates over all newly processed viewpoints between time $t$ and $t+1$.
The term $\Delta_{v} = t_{v} - t_{v+1}$ expresses the time difference between two consecutive viewpoints.
This formulation allows online computation during the flight without extra memory requirements.

\subsection{The system matrix}
\label{sec:system_matrix}
The system matrix represents the projection of Compton cones to $\mathcal{M}$.
The elements $t_{ij}$ are computed as:
\begin{equation}
      t_{ij} = e^{-(\mu \lVert \mathbf{m}_{j} - \mathbf{s}_{i} \rVert)}\ 
     \mathrm{h}(\delta_{ij})
  \frac{\mathrm{L}(\phi_{ij}, \theta_{ij})}{\lVert \mathbf{m}_{j} - \mathbf{s}_{i} \rVert ^ {2} },
\end{equation}
where 
$\mu $ is the linear attenuation coefficient of the environment ($\approx \SI{0.01}{\meter^{-1}} $ for $\SI{622}{\kilo\electronvolt}$ photons), 
$\phi_{ij}, \theta_{ij}$ are the polar coordinates determining the relative position of $\mathbf{m}_{j}$ and the sensor position $\mathbf{\hat{s}}_{i}$,
$h(\delta_{ij})$ is the projection function, and 
$\delta_{ij}$ is the angle difference (see \reffig{fig:setup}). 
The projection function
\begin{equation}
\mathrm{h}(\delta_{ij}) = e^{-\frac{1}{2} \left (  \frac{\delta_{ij}}{\sigma}    \right )^{2}}
\end{equation}

projects the cone to $\mathcal{M}$ and models uncertainty in the Compton angle estimation.

\section{MULTIROBOT SEARCH STRATEGY}
\label{sec:search_strategy}

An important part of the proposed radiation mapping method is a novel active search strategy for a group of autonomous \acp{MAV} equipped with Compton cameras.
The proposed strategy plans the future movement of the \acp{MAV} to increase the information gain and take advantage of the \acp{MAV}' mobility.
We assume that the communication between \acp{MAV} is available and utilize a centralized approach, where one drone plans the future actions for all \acp{MAV}.
We are aware that many different strategies might be designed based on different robotic missions or required optimality criteria.
In this paper, we focus on exploiting the advantages of direction-based radiation sensors in practice and present an efficient solution to the autonomous search for radiation sources.
Previous works \cite{baca_gamma_2021_icuas, mascarich_autonomous_2021_ICRA} show that an active search strategy is beneficial compared to extensive exploration of the whole area following a predefined path. Further, an online method can speed up the search process, find the sources faster, and compensate for the effect of less accurate measurements, noise, and background radiation.
The task can be divided into two subtasks: 
\textbf{exploration} --- visiting areas where no measurements have been collected yet and where an undiscovered ionizing radiation source might be present, and \textbf{exploitation} --- flying closer to the previously estimated source position to either collect more measurements and make the localization more precise, or to disprove the presence of a radiation source in that location. 

\begin{figure}[!h]
  \centering
  \includegraphics[width=0.5\textwidth,trim={0cm 2.5cm 0cm 0cm},clip]{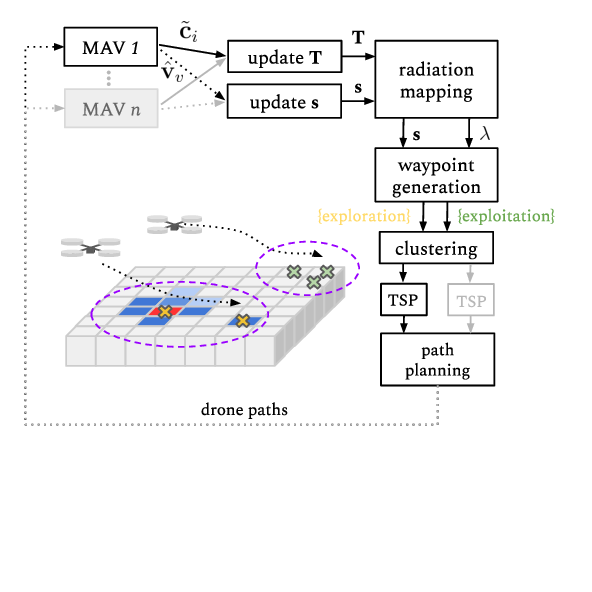}
  \caption{The search strategy pipeline. The future waypoints for the \acp{MAV} are generated based on $\mathbf{s}$ (exploration --- green) and $\bm{\lambda}$ (exploitation --- orange). The waypoints are assigned to individual \acp{MAV} by dividing them into clusters based on their spatial position (purple circles). The ordering of waypoints is determined using the sequencing method (TSP). Finally, the pipeline outputs non-colliding paths for every \ac{UAV}. }
  \label{fig:system_scheme}
\end{figure}

The search strategy is illustrated in \reffig{fig:system_scheme}.
The \textbf{waypoint generation} method generates waypoints based on the most recent estimate of $\bm{\lambda}$ and $\mathbf{s}$, as shown in \refalg{alg:wg}.
The point $\mathbf{m}_{j}$ is taken as the \textit{exploration} waypoint if $\mathbf{s}_{j} < s_{min}$, where $s_{min}$ is a preset threshold specifying the minimum required chance that a particle emitted at this position has been detected by the \acp{MAV}.
The \textit{exploration} waypoints $\mathbf{w}_{\mathbf{s}}$ are then clustered using the K-Means method to reduce their number.
For the \textit{exploitation} waypoints $\mathbf{w}_{\bm{\lambda}}$, we use local maxima in the map $\mathcal{M}$ based on the latest $\bm{\lambda}$.

 \begin{algorithm}
    \scriptsize
   \caption{Waypoint generation}\label{alg:wg}
   \begin{algorithmic}[1]
     \State \textbf{Input:}  $\bm{\lambda}$, $\mathbf{s}$
     \Procedure {generate\_waypoints}{$\bm{\lambda}, \mathbf{s}$} 

     \State $\mathbf{w}_{\bm{\lambda}} \gets \{\}$ \Comment{Exploitation waypoints}
     \State $\mathbf{w}_{\mathbf{s}} \gets \{\}$ \Comment{Exploration waypoints}
     \For{$\mathbf{m}_{j}$ in $\mathcal{M}$}{}
       \If{$\bm{\lambda}_{j}>\bm{\lambda}_{k}, \forall \ \mathbf{m}_k \in neighbourhood(\mathbf{m}_{j})$}
          \If{$\mathbf{s}_{j}<s_{max}$} 
                \State $\mathbf{w}_{\bm{\lambda}} \gets \mathbf{w}_{\bm{\lambda}} \cup \{\mathbf{m}_{j}\}$
          \EndIf
       \EndIf
       \If{$\mathbf{s}_{j}<s_{min}$}
          \State $\mathbf{w}_{\mathbf{s}} \gets \mathbf{w}_{\mathbf{s}} \cup \{\mathbf{m}_{j}\}$
        \EndIf 
       \State $\mathbf{w}_{\mathbf{s}} \gets \mathrm{cluster}(\mathbf{w}_{\mathbf{s}})$ \Comment{K-means}
        \State $\mathbf{w} \gets \mathrm{filter\_recent}(\mathbf{w}_{\bm{\lambda}} \cup \mathbf{w}_{\bm{s}})$ \Comment{filter visited}
     \EndFor
     \EndProcedure
     \State \textbf{Output:} $\mathbf{w}$
   \end{algorithmic}
   
 \end{algorithm}

While the waypoints that have not recently been visited by any of the \ac{MAV} are prioritized, the generated waypoints $\{\mathbf{w}_{\bm{\lambda}} \cup \mathbf{w}_{\mathbf{s}}\}$ are filtered and assigned to individual \acp{MAV} using a \textbf{clustering} method.
The number of clusters equals the number of \acp{MAV}, and each cluster centroid is initialized at the current position of each \ac{MAV}. 
The waypoints assigned to each \ac{MAV} are then ordered into an optimal sequence using a \ac{TSP} solver, where the sequence always starts at the current position of the \ac{MAV} and ends in any of the assigned waypoints.
Finally, the \textbf{path planning} method plans non-colliding paths for all \acp{MAV}, visiting all assigned waypoints using the A* algorithm.

\section{EXPERIMENTS}

The proposed radiation mapping method combined with the active search strategy was tested in simulated and real-world experiments.
The number of iterations used in \ac{MLEM} was empirically set to 10, as it represents a good balance between contrast recovery and image noise amplification.
The map resolution $r$ was set to $\SI{0.5}{\meter}$ during all the presented experiments.

\subsection{Simulation}

\begin{figure}[h!tb]
    \begin{subfigure}{0.24\textwidth}
        \includegraphics[width=\textwidth, trim={1cm 1.7cm 0.5cm 2.5cm}, clip]{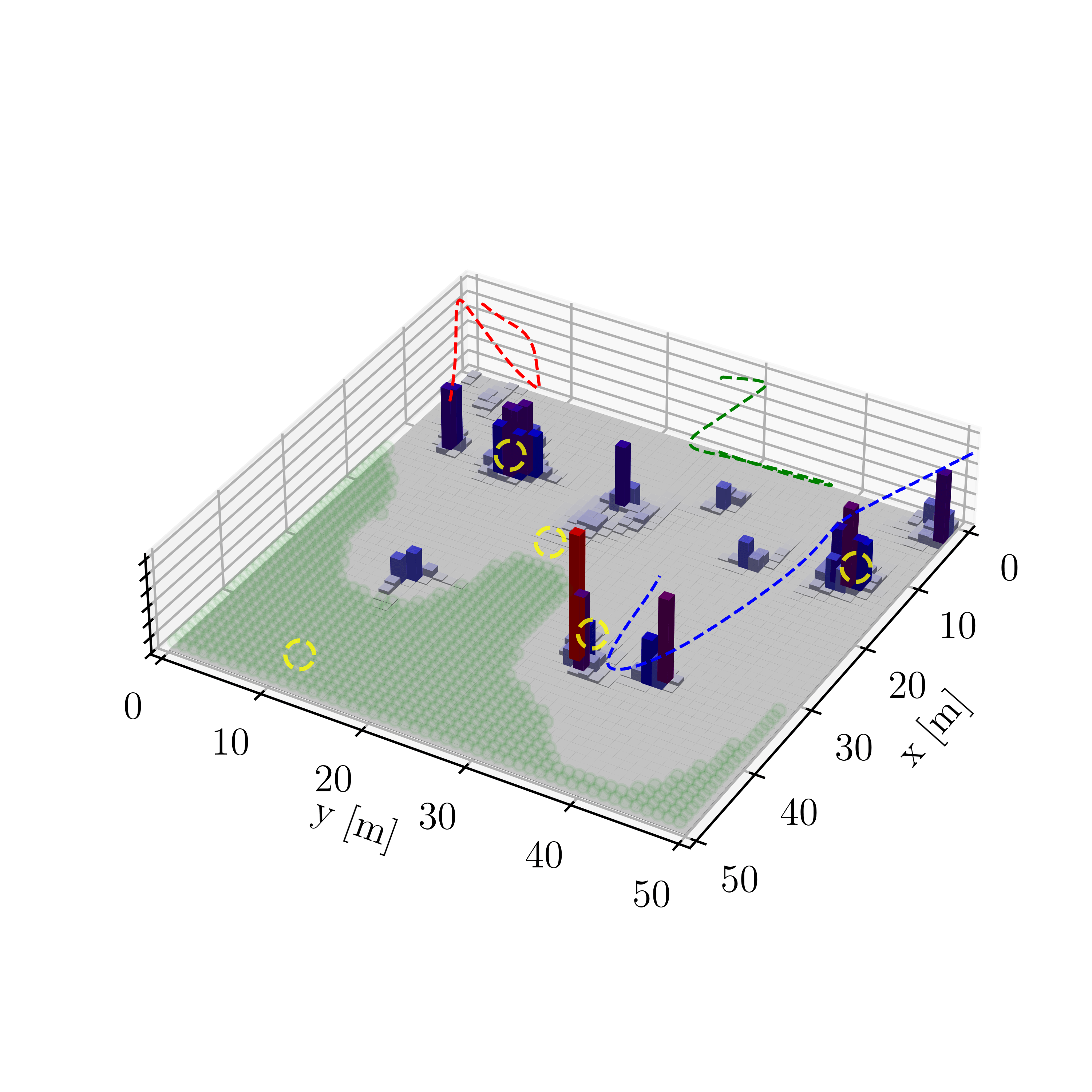}
        \caption{$t =\SI{15}{\second}$}
        \label{}
    \end{subfigure} 
    \begin{subfigure}{0.24\textwidth}
        \includegraphics[width=\textwidth, trim={1cm 1.7cm 0.5cm 2.5cm}, clip]{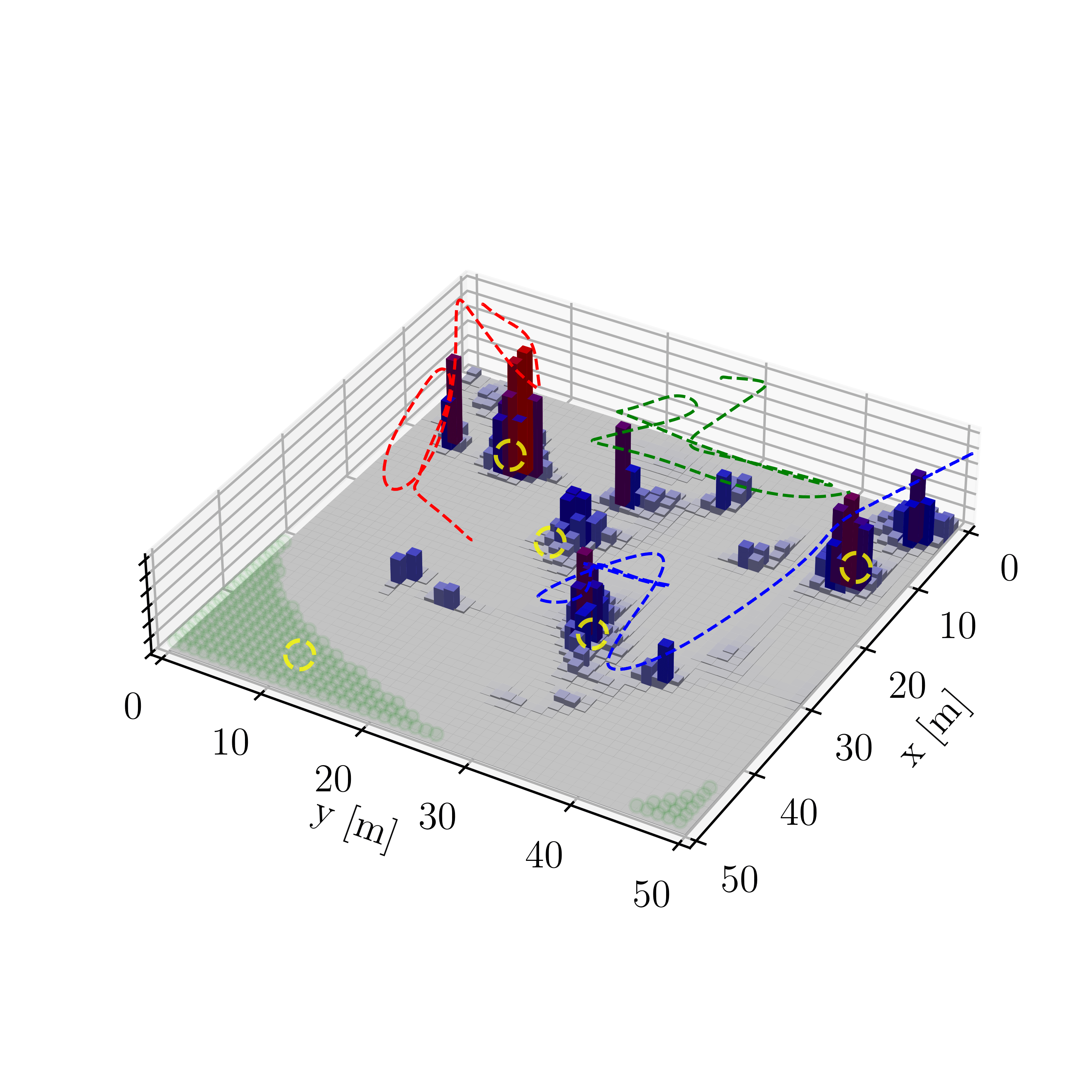}
        \caption{$t =\SI{30}{\second}$}
        \label{}
    \end{subfigure}
    \begin{subfigure}{0.24\textwidth}
        \includegraphics[width=\textwidth, trim={1cm 1.7cm 0.5cm 2.5cm}, clip]{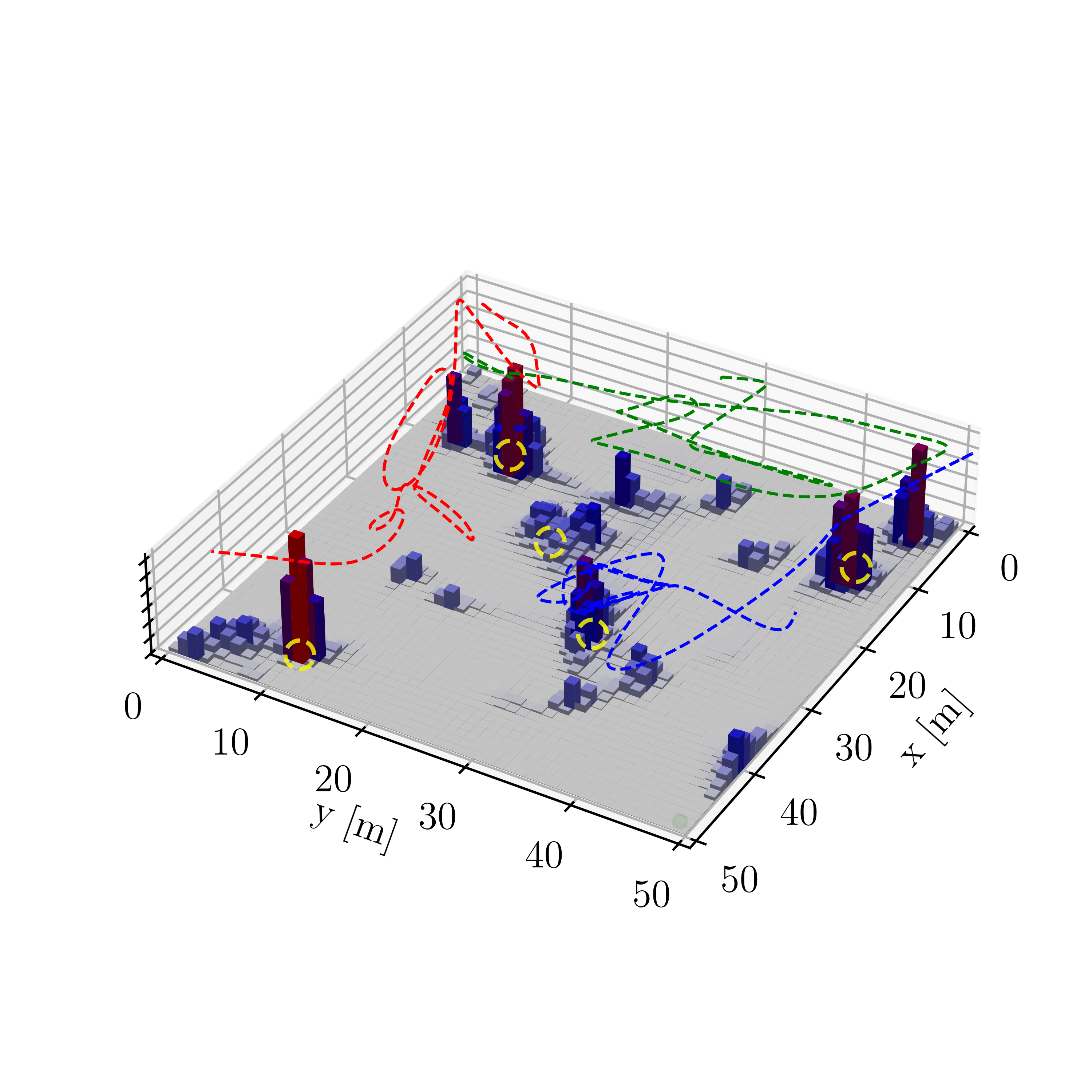}
        \caption{$t =\SI{45}{\second}$}
        \label{}
    \end{subfigure}
    \begin{subfigure}{0.24\textwidth}
        \includegraphics[width=\textwidth, trim={1cm 1.7cm 0.5cm 2.5cm}, clip]{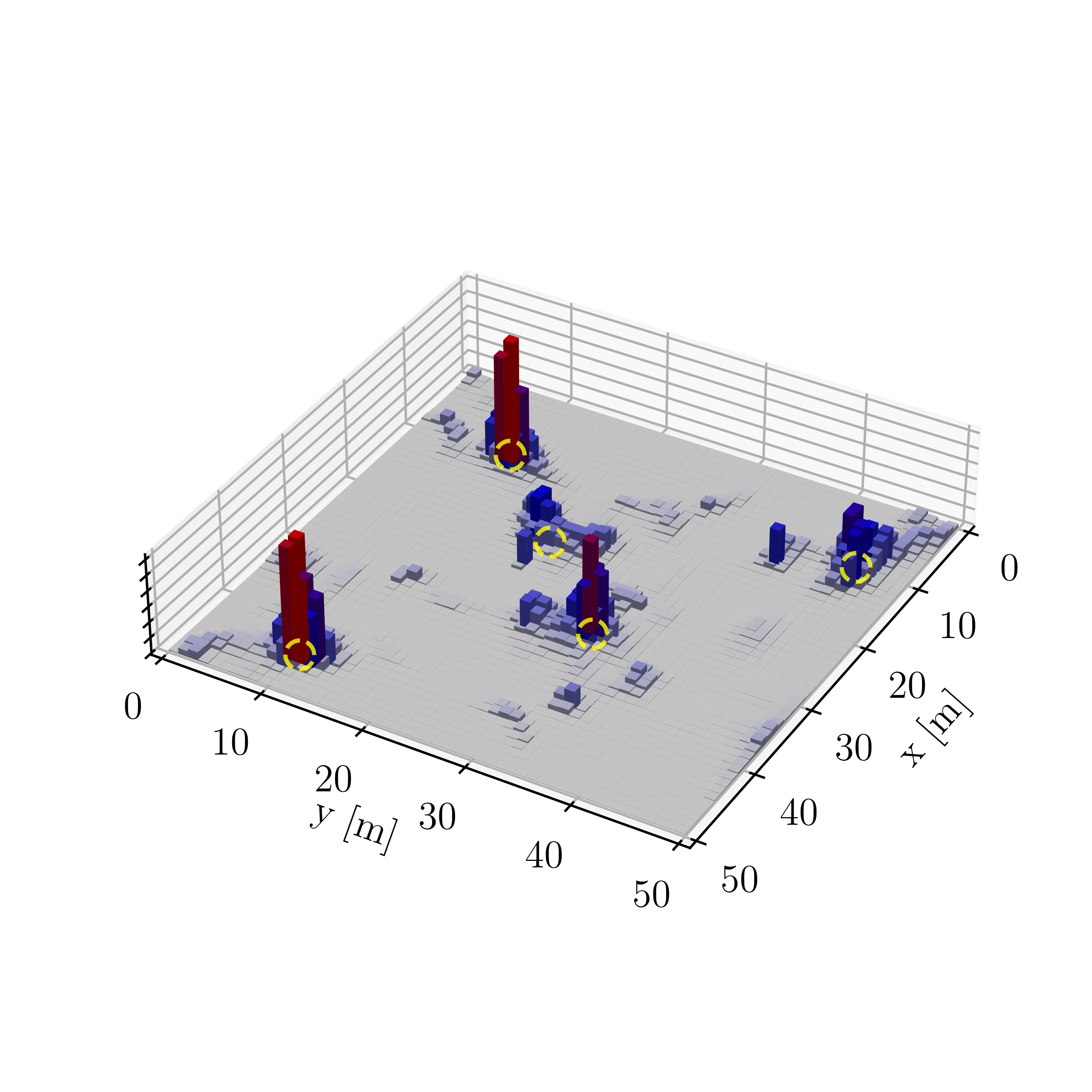}
        \caption{$t =\SI{100}{\second}$}
        \label{}
    \end{subfigure}
    \caption{The progress of one simulated search mission with three \acp{MAV}. The estimated emission intensity $\bm{\lambda}$ is shown as the 3D bars (grey-blue-red colour scale). The unexplored area ($s_{j} < s_{min}$) is shown in light green, and the \acp{MAV}' trajectories are depicted as dashed lines. The ground truth positions of the $\SI{2}{\giga \becquerel}$ sources are shown with yellow dashed circles.}
    \label{fig:sim_progress}
    \vspace{-0.5cm}
\end{figure}

\begin{figure}[h!]
    \includegraphics[width=0.5\textwidth, trim={0.4cm 0.5cm 0.0cm 0.6cm}, clip]{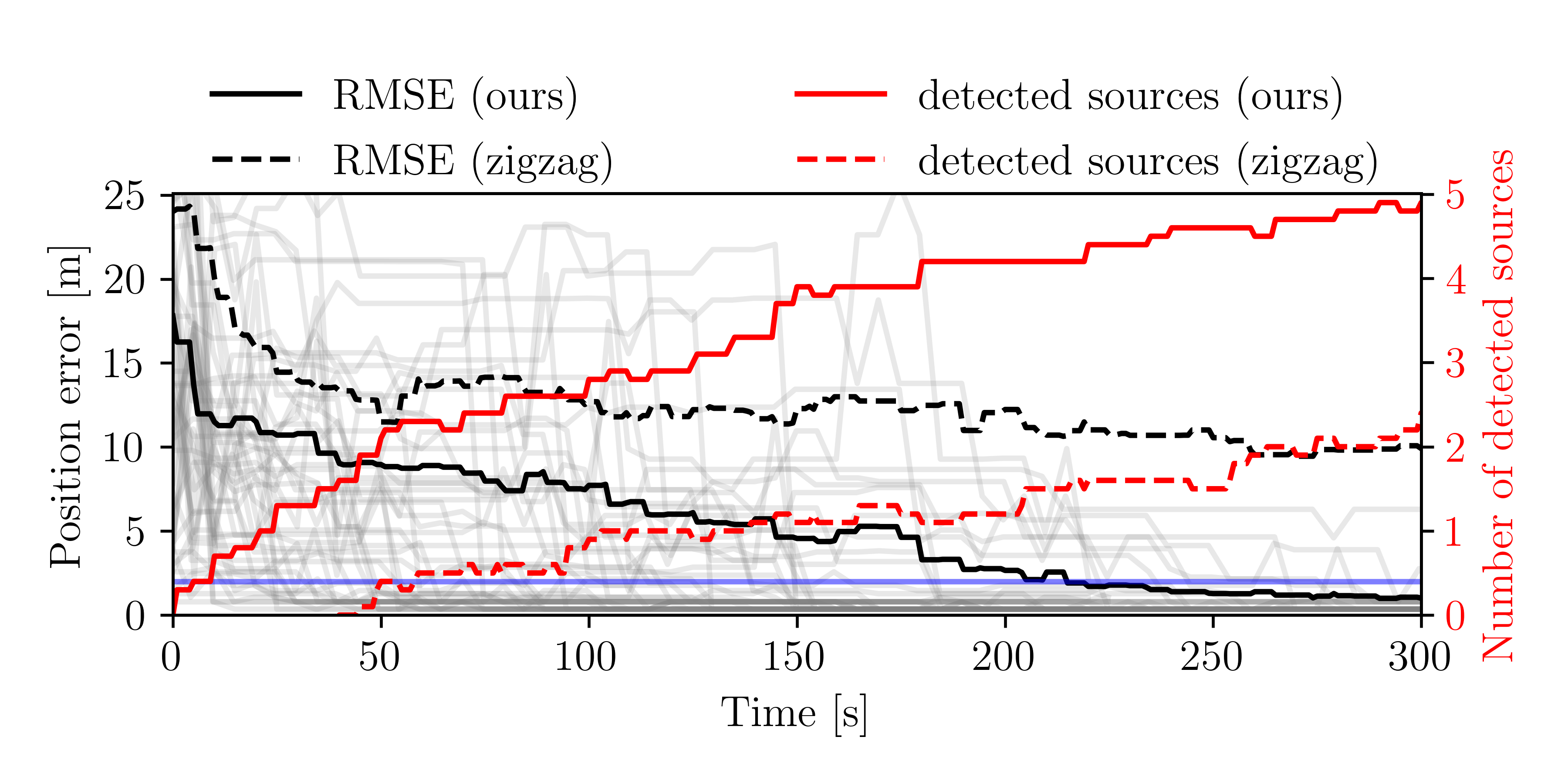}     
    \caption{The localization error for each simulated source for 10 simulated experiments (light grey), RMSE (black) and the average number of the localized sources within \SI{2}{\meter} precision (red). We compare the proposed search strategy (solid line) with a predefined zigzag search pattern (dashed line).  
    }
    \vspace{-0.5cm}
    \label{fig:sim_stats}
\end{figure}

Three \acp{MAV} equipped with simulated \ac{pix} Compton cameras were used in the experiment.
The maximum speed of the \ac{MAV} was set to $\SI{8}{\meter \per \second}$ with the height of flight at $\SI{2}{\meter}$ above ground.
There were five simulated sources, each with an emission activity $\SI{2}{\giga \becquerel}$, located within the $50 \times 50 \ \si{\meter}$ area, as shown in \reffig{fig:sim_progress}.
The simulated experiments were performed in the realistic simulator (the MRS multirotor simulator \cite{mrs_system}).
The properties of simulated ionizing radiation and the \ac{pix} Compton camera are described in \cite{baca_2019_timepix_iros}.
The progress of one simulated experiment is shown in \reffig{fig:sim_progress}.
The \acp{MAV} simultaneously explored the unexplored area (green) and,  by collecting additional measurements, improved the quality of reconstruction based on previous measurements.
The same scenario was repeated 10 times with results shown in \reffig{fig:sim_stats}.
For the statistical evaluation, we use the following metric where we take five local maxima (equaling the number of radiation sources in the area) with the highest $\lambda_{j}$ value as the estimated source positions. 
The recorded localization error is the distance between the ground truth position of the radiation source and the nearest estimated source position. 
As shown in \reffig{fig:sim_stats}, the proposed search strategy converges to the true source positions, with the \ac{RMSE} decreasing below $\SI{2}{\meter}$ precision threshold after $\SI{220}{\second}$. All simulated sources are localized within $\SI{2}{\meter}$ precision after $\SI{300}{\second}$.
The effectiveness of the presented search strategy is demonstrated by comparing it with a predefined zigzag search pattern (also repeated 10 times), where the search area is divided into three equal parts, and each part is covered by an \ac{MAV} following a zigzag trajectory with a $\SI{2}{\meter}$ lateral step and a maximum flight speed of $\SI{8}{\meter \per \second}$. 
In comparison, the zigzag search strategy shows slower convergence than our approach because the \acp{MAV} following predefined paths do not fly closer to the estimated source positions, resulting in fewer collected measurements and less accurate reconstruction of the radiation sources.

\subsection{Real-world experiment}
\begin{figure}[h!]
    \begin{subfigure}{0.24\textwidth}
        \includegraphics[width=\textwidth, trim={0cm 0cm 0cm 0cm}, clip]{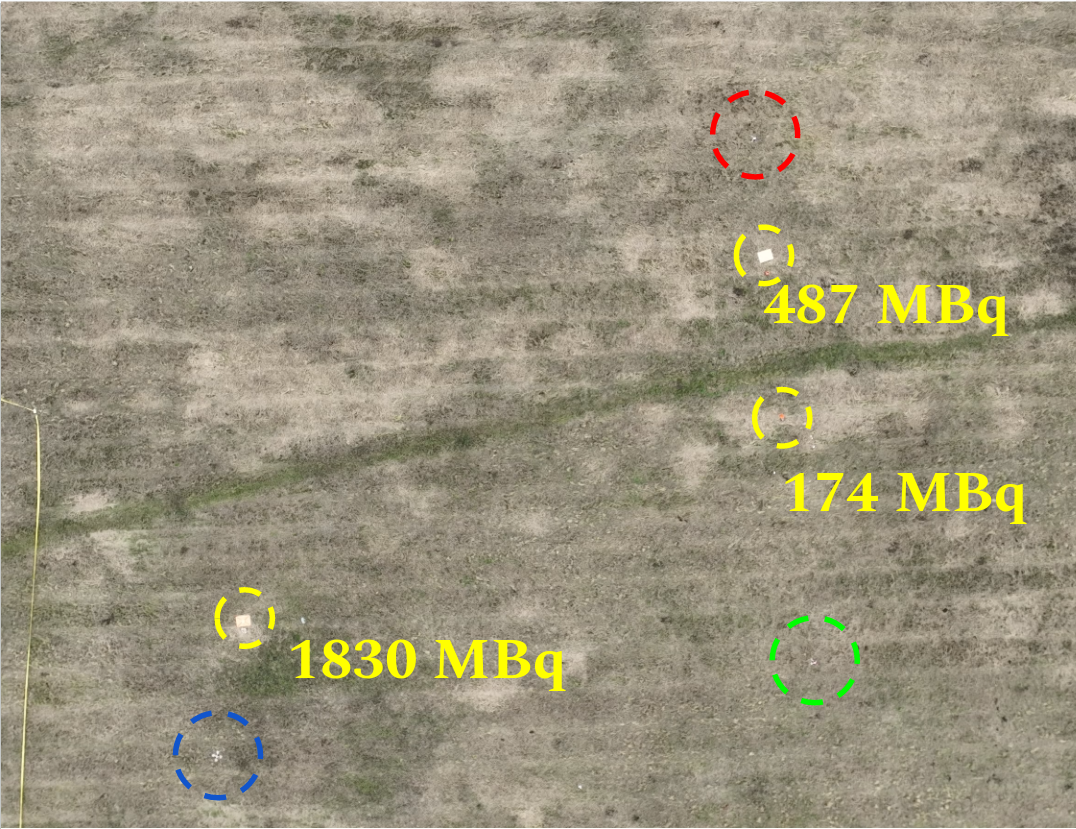}
        \caption{The \acp{UAV} (\red{red}/\green{green}/\blue{blue}) and radiation sources (yellow) during the experiment ($t = \SI{32}{\second}$).}
        \label{fig_a}
    \end{subfigure}
    \begin{subfigure}{0.24\textwidth}
        \includegraphics[width=\textwidth, trim={0cm 0cm 0cm 0cm}, clip]{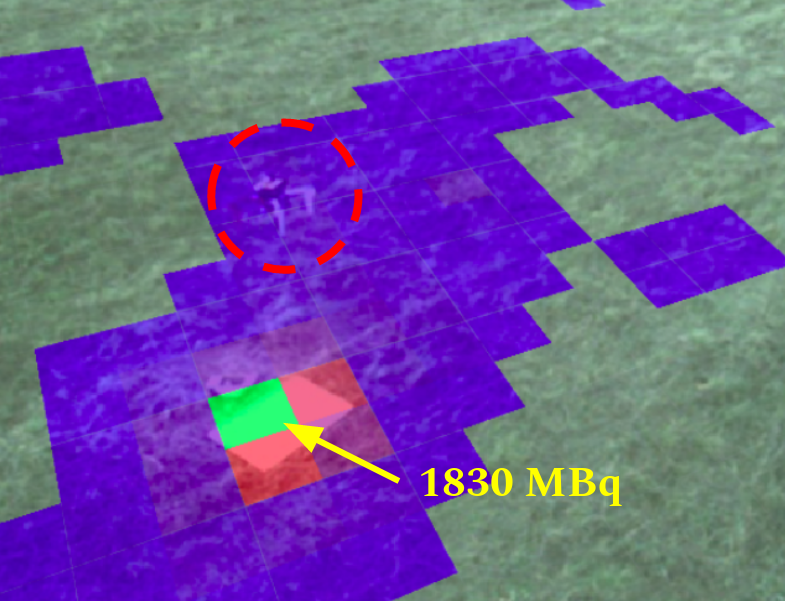}
        \caption{Estimated radiation intensity around the \SI{1830}{\mega\becquerel} radiation source with \ac{UAV} above. ($t = \SI{520}{\second}$)}
        \label{}
    \end{subfigure}
    \caption{The real-world experiment with Cs-137 radiation sources.}
    \label{fig:pole}
    \vspace{-0.3cm}
\end{figure}

Conducting real-world experiments with real ionizing radiation sources is complicated due to logistical reasons requiring coordination between multiple institutions, as well as legal restrictions.
Consequently, the real-world experiments had to be performed within a short window of several hours.
We present the real-world experiment as a proof of concept and validation of the proposed method.
The system was tested in an open field $50 \times 70 \ \si{\meter}$ where three Cesium-137 sources with emission activity $1830, 487, 174\ \si{\mega\becquerel}$ were present, as shown in \reffig{fig:pole}.
We demonstrated the proposed radiation mapping method using three F450 \acp{UAV} \cite{hert_2023_jint}, each of which was equipped with an Intel NUC onboard computer, a single \ac{pix} Compton camera, a wireless communication module, and a \ac{RTK} module for localization of the \acp{UAV}.
The horizontal speed constraints of the platform were set to $\SI{4}{\meter \per \second}$, and the estimation method was running in real time.
However, an error in the drone localization method appeared during the flight (not related to this paper's content), so the presented results had to be recomputed later on recorded rosbag data.

\begin{figure}[h!]
    \begin{subfigure}{0.23\textwidth}
        \includegraphics[width=\textwidth, trim={1cm 2.1cm 0.5cm 3cm}, clip]{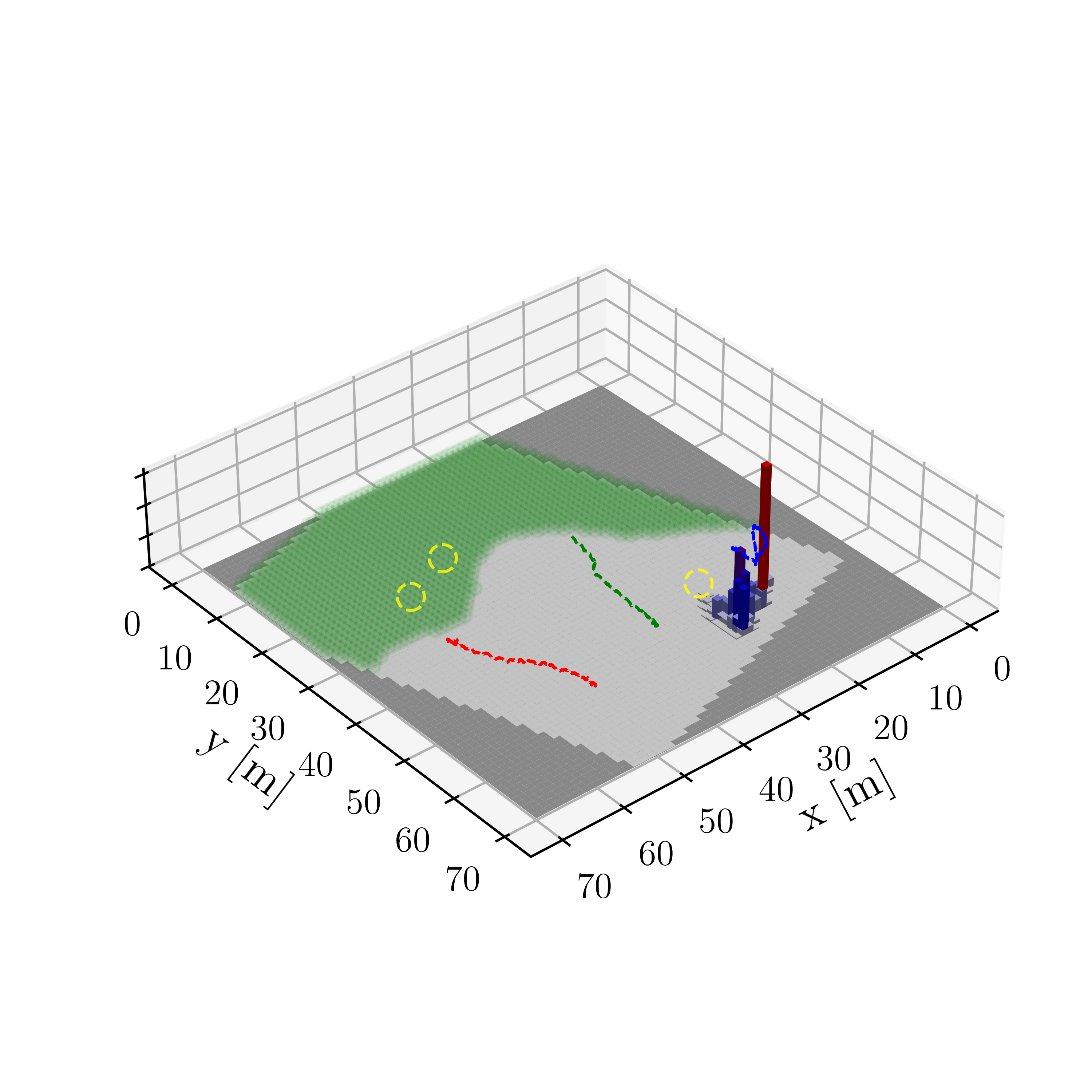}
        \caption{$t =\SI{10}{\second}$}
        \label{}
    \end{subfigure}
    \begin{subfigure}{0.23\textwidth}
        \includegraphics[width=\textwidth, trim={1cm 2.1cm 0.5cm 3cm}, clip]{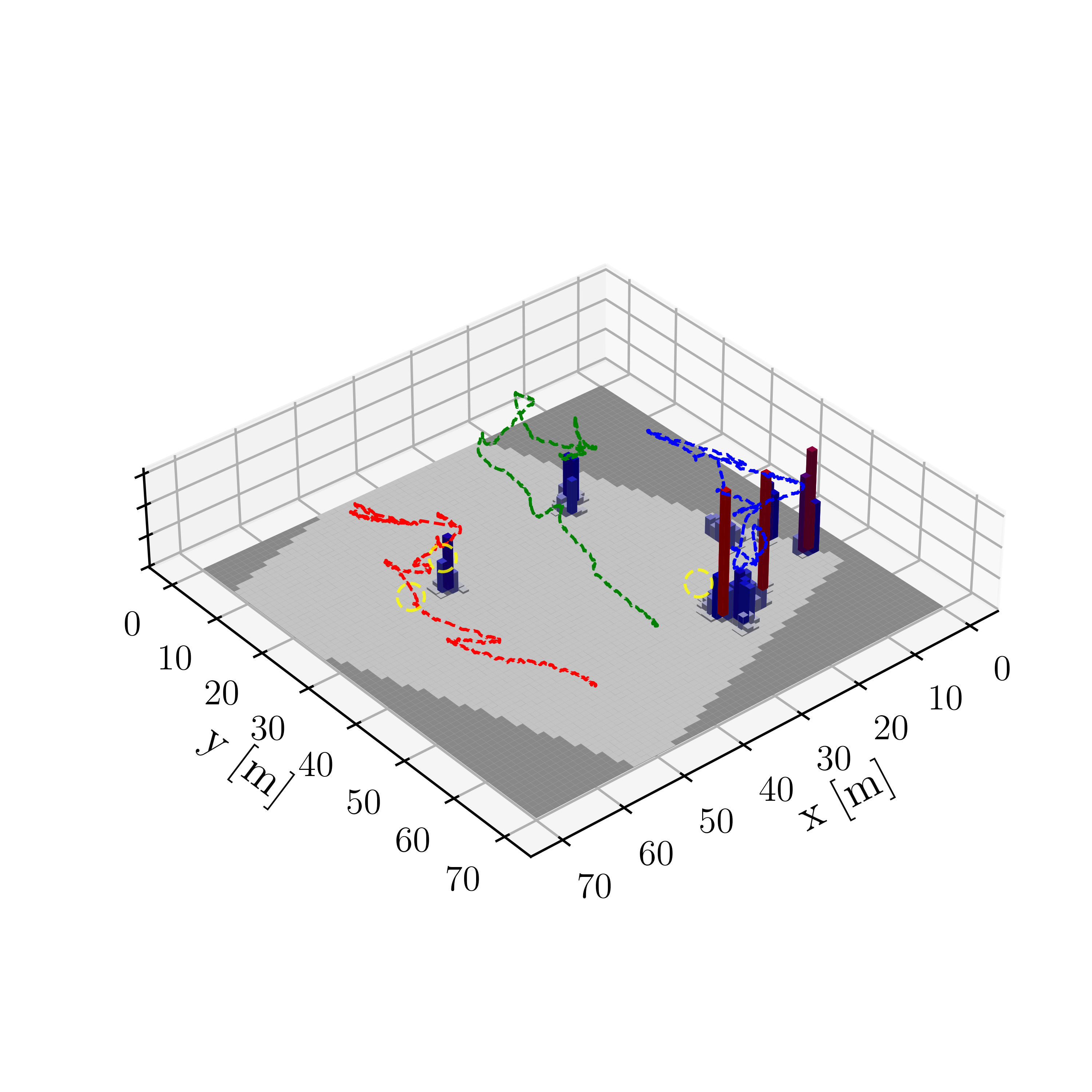}
        \caption{$t =\SI{81}{\second}$}
        \label{}
    \end{subfigure}
    \begin{subfigure}{0.23\textwidth}
        \includegraphics[width=\textwidth, trim={1cm 2.1cm 0.5cm 3cm}, clip]{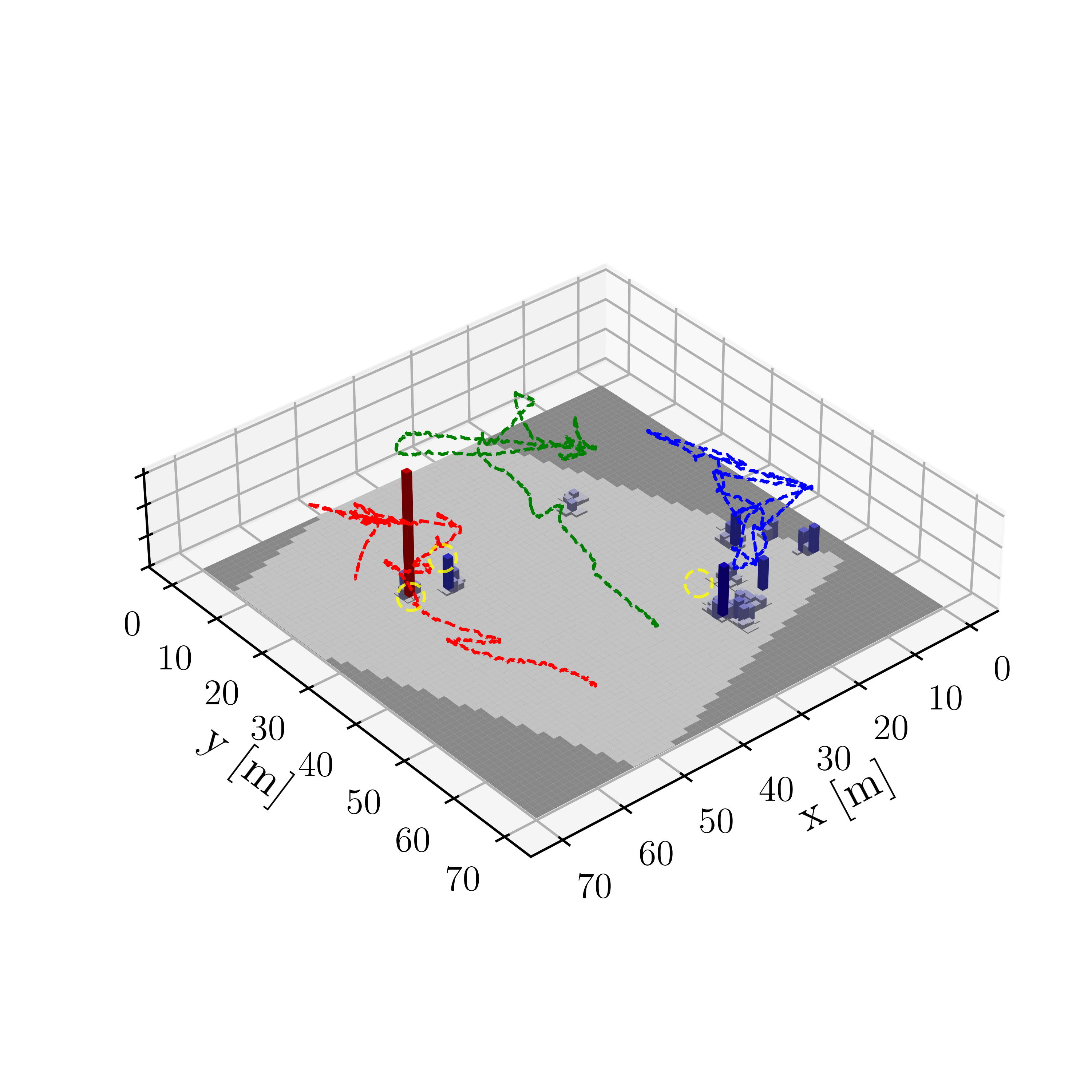}
        \caption{$t =\SI{108}{\second}$}
        \label{}
    \end{subfigure}
    \centering
    \begin{subfigure}{0.23\textwidth}
        \includegraphics[width=\textwidth, trim={1cm 2.1cm 0.5cm 3cm}, clip]{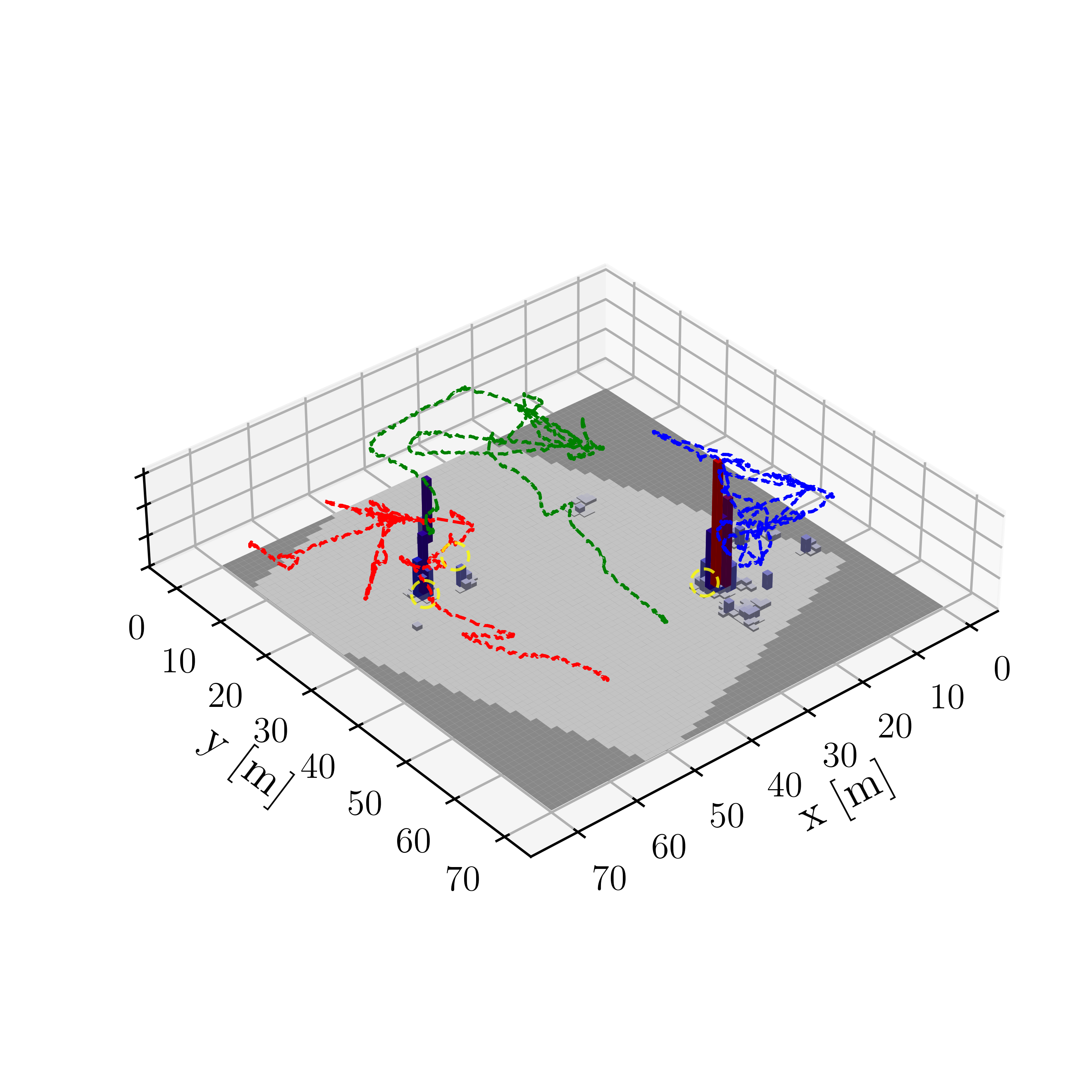}
        \caption{$t =\SI{135}{\second}$}
        \label{}
    \end{subfigure}
    \begin{subfigure}{0.49\textwidth}
        \includegraphics[width=\textwidth, trim={1cm 2cm 0.5cm 3cm}, clip]{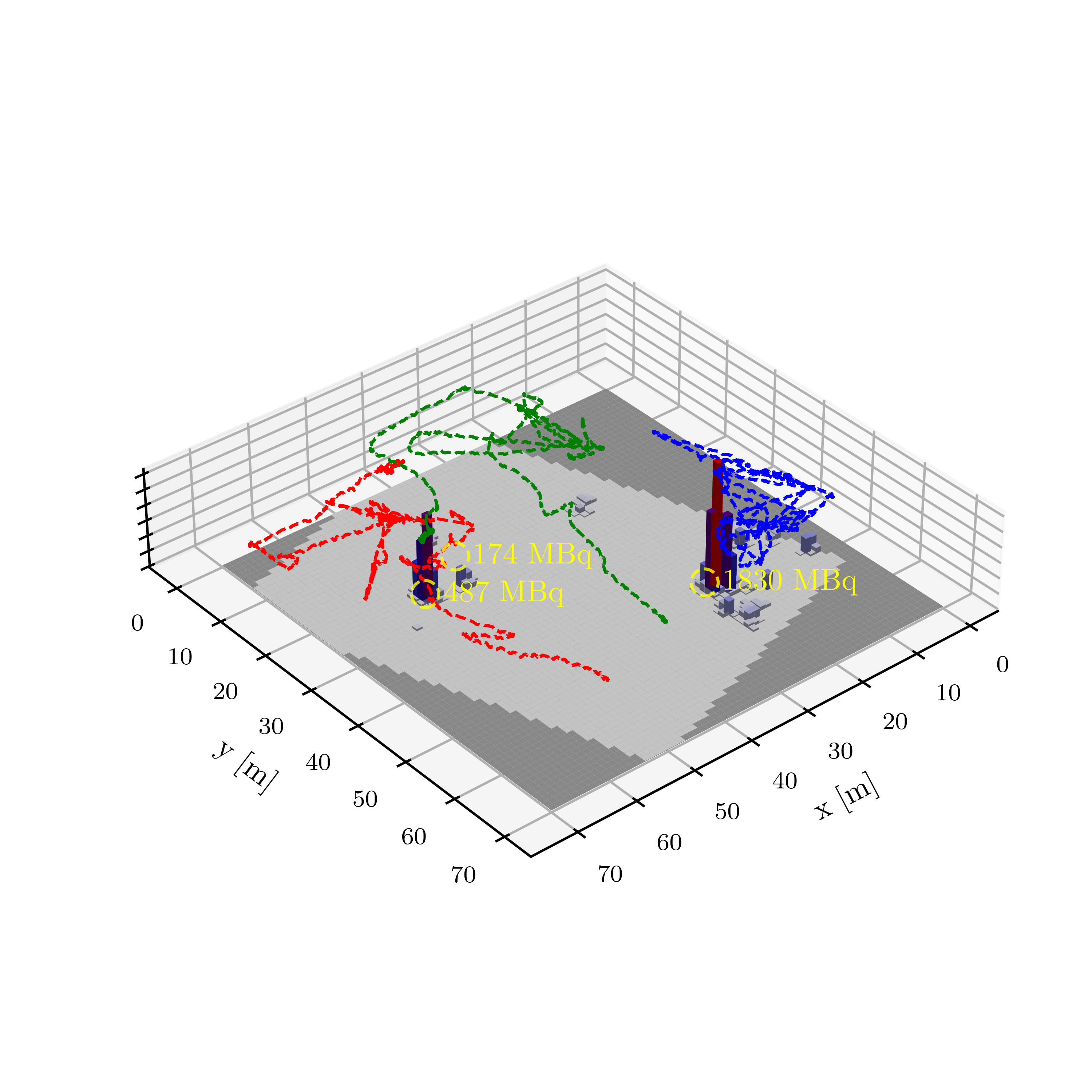}
        \caption{$t =\SI{240}{\second}$}
        \label{}
    \end{subfigure}
    \caption{Progress of the real-world search mission with three \acp{UAV}. The estimated emission intensity $\bm{\lambda}$ is shown as the 3D bars (grey-blue-red colour scale), the unexplored area ($s_{j} < s_{min}$) is shown in light green, and the \acp{MAV} trajectories are depicted as dashed lines. The ground truth positions of the radiation sources are shown with yellow dashed circles.}
    \label{fig:last_flight_results}
    \vspace{-0.7cm}
\end{figure}

The progress of the radiation mapping is shown in \reffig{fig:last_flight_results}.
Multiple Compton cones caused by the $\SI{1830}{\mega \becquerel}$ radiation source were recorded right after the start of the experiment.
However, the \ac{UAV} had to collect measurements from multiple positions in order to precisely localize the source position, possibly due to outliers, measurement noise, or background radiation. 
Two of three Cesium-137 were localized within $\SI{108}{\second}$.
After collecting more measurements, the estimates converged to their true position while correctly evaluating their relative emission activity.
The third least active $\SI{174}{\mega \becquerel}$ source remained undiscovered due to its low emission activity, which shows the detection limits of the proposed system.
The statistics of the recorded radiation data are shown in \reftab{tab:last_flight_stats}.
The video from the experiment is available at \href{https://mrs.felk.cvut.cz/iros-compton-2024}{https://mrs.felk.cvut.cz/iros-compton-2024}.

\begin{figure}[h!]
    \includegraphics[width=0.5\textwidth, trim={0.3cm 0.3cm 0cm 0.4cm}, clip]{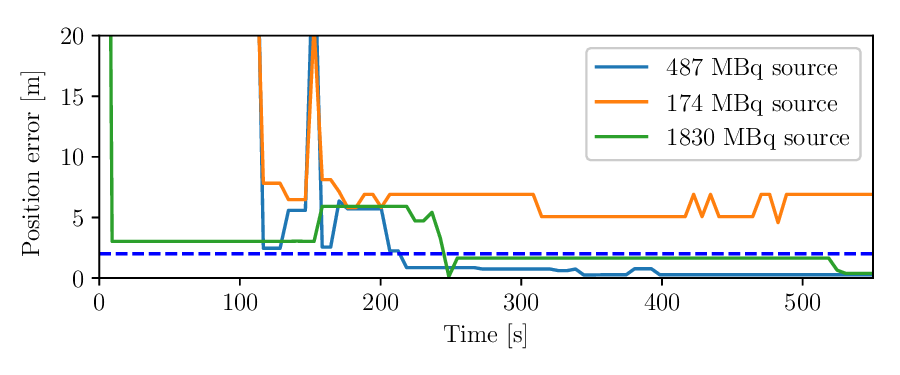}     
    \caption{Localization error during the real-world experiment. Both \SI{1830}{\mega\becquerel} and \SI{487}{\mega\becquerel} were localized within \SI{5}{\meter} precision after \SI{120}{\second}. The method further converged during the rest of the experiment. The least active source was not precisely localized.}
    \label{fig:last_flight_stats}
\end{figure}

The experiment demonstrates the ability to localize multiple radiation sources of different emission activity in a large, open-space area.
The reconstruction method correctly localized two out of three Cesium-137 sources within $\SI{1}{\meter}$ precision, as shown in \reffig{fig:last_flight_stats}.
The proposed method should be further tested with additional (or even distributed) radiation sources and with more agile \ac{MAV} capable of flying up to $\SI{8}{\meter \per \second}$ or more, as these were unavailable at the time of experimentation.

\begin{table}[h]
\label{table_example}
\begin{center}
\begin{tabular}{|c|c|c|c|}
\hline
Event type & Count & Rate [\si{\per \second}]& Relative share \\
\hline 
Photoelectric & 1503 & 10.02& \SI{93.4}{\percent}  \\
\hline
Compton e. (cones)& 69 (78) & 0.46 (0.52) & \SI{4.3}{\percent}\\
\hline
Background & 38 & 0.25 & \SI{2.3}{\percent}\\
\hline
\end{tabular}
\end{center}
\caption{Radiation measurements statistics for the real-world experiment (for $t < \SI{240}{\second}$). We denote the interactions with measured energy $>\SI{800}{\kilo \electronvolt}$ as background, as such particles are too energetic to be emitted by the Cesium-137 sources. Notably, only $\approx 4.2 \%$ of recorded events resulted in the Compton measurement.}
\label{tab:last_flight_stats}
\vspace{-0.5pt}
\end{table}

\section{CONCLUSION}

In this work, we have presented a novel approach to the autonomous localization of multiple ionizing radiation sources. This was done by using single-layer Compton cameras onboard \acp{MAV} working in a team.
We have enabled online radiation mapping by modeling the detection sensitivity of the Compton camera, and the reconstructive projection of Compton cones into a volumetric map.
Furthermore, we have expanded the capabilities of the maximum likelihood radiation detection method by coupling it with an active search strategy.
This strategy leverages the online processing of Compton measurements to plan the future movements of the \acp{MAV}. 
The method has been validated in both simulation and a real-world experiment, demonstrating the system's ability to autonomously detect previously unknown radiation sources, and to improve the quality of source localization during flight.

The proposed method relies heavily on the occurrence of the Compton scattering effect.
The data collected from the real-world experiment highlighted that this effect comprises only a small subset of all radiation events captured by the detector.
A potential direction for future work involves combining the direction-based Compton camera approach with intensity-based methods.
Developing estimation methods that merge the advantages of both could significantly enhance the robots' capabilities in terms of fast and accurate localization of harmful ionizing radiation sources.




\section*{ACKNOWLEDGMENTS}
This work was funded by CTU grant no SGS23/177/OHK3/3T/13, by the Czech Science Foundation (GA\v{C}R) under research project no. 23-07517S and by the European Union under the project Robotics and advanced industrial production (reg. no. CZ.02.01.01/00/22\_008/0004590).

\bibliographystyle{IEEEtran}
\bibliography{root}

\end{document}